\def\paperID{10655}
\def\confName{ICCV}
\def\confYear{2025}

\def\paperTitle{OpenInsGaussian: Open-vocabulary Instance Gaussian Segmentation with Context-aware Cross-view Fusion}

\def\authorBlock{
    Tianyu Huang$^{1}$\quad
    Runnan Chen$^{1}$\quad
    Dongting Hu$^{2}$\quad
    Fengming Huang$^{1}$\quad
    Mingming Gong$^{2}$\quad
    Tongliang Liu$^{1}$\\[0.3em]
    $^{1}$University of Sydney\qquad
    $^{2}$University of Melbourne\\[0.3em]
}

\newif\ifreview 
\newif\ifarxiv \newcommand{\arxiv}{\arxivtrue}
\newif\ifcamera 
\newif\ifrebuttal 

\arxiv 

\pdfoutput=1
\documentclass[10pt,twocolumn,letterpaper]{article}
\ifreview \usepackage[review]{cvpr} \fi
\ifarxiv \usepackage[pagenumbers]{cvpr} \fi
\ifrebuttal \usepackage[rebuttal]{cvpr} \fi
\ifcamera \usepackage{cvpr} \fi

\usepackage{graphicx}	
\usepackage{amsmath}	
\usepackage{amssymb}	
\usepackage{booktabs}
\usepackage{times}
\usepackage{microtype}
\usepackage{epsfig}
\usepackage{caption}
\usepackage{float}
\usepackage{placeins}
\usepackage{color, colortbl}
\usepackage{stfloats}
\usepackage{enumitem}
\usepackage{tabularx}
\usepackage{xstring}
\usepackage{multirow}
\usepackage{xspace}
\usepackage{url}
\usepackage{subcaption}
\usepackage{xcolor}
\usepackage[hang,flushmargin]{footmisc}

\ifcamera \usepackage[accsupp]{axessibility} \fi

% ...

\ifarxiv  \fi

\newcommand{\R}[1]{{%
    \textbf{%
        \ifstrequal{#1}{1}{\textcolor{red}{R#1}}{%
        \ifstrequal{#1}{2}{\textcolor{blue}{R#1}}{%
        \ifstrequal{#1}{3}{\textcolor{magenta}{R#1}}{%
        \ifstrequal{#1}{4}{\textcolor{teal}{R#1}}{%
                           \textcolor{cyan}{R#1}%
        }}}}%
    }%
}}

\usepackage{xr-hyper}

\makeatletter
\newcommand*{\addFileDependency}[1]{
  \typeout{(#1)}
  \@addtofilelist{#1}
  \IfFileExists{#1}{}{\typeout{No file #1.}}
}

\makeatother
\newcommand*{\myexternaldocument}[1]{
    \externaldocument{#1}
    \addFileDependency{#1.tex}
    \addFileDependency{#1.aux}
}

\definecolor{cvprblue}{rgb}{0.21,0.49,0.74}
\usepackage[pagebackref,breaklinks,colorlinks,allcolors=cvprblue]{hyperref}
\usepackage[capitalize]{cleveref}
\crefname{section}{Sec.}{Secs.}
\crefname{table}{Table}{Tables}
\crefname{figure}{Fig.}{Figs.}

\ifarxiv \crefname{appendix}{App.}{Apps.}
\else \crefname{appendix}{Suppl.}{Suppls.} \fi

\usepackage[accsupp]{axessibility}  

\usepackage{comment}

\unless\ifarxiv \myexternaldocument{_supplementary} \fi
\title{\paperTitle}
\author{\authorBlock}

\begin{document}
\twocolumn[{
    \renewcommand\twocolumn[1][]{#1}
    \maketitle
    \centering
    \vspace{-0.6cm}
    \includegraphics[width=0.98\textwidth]{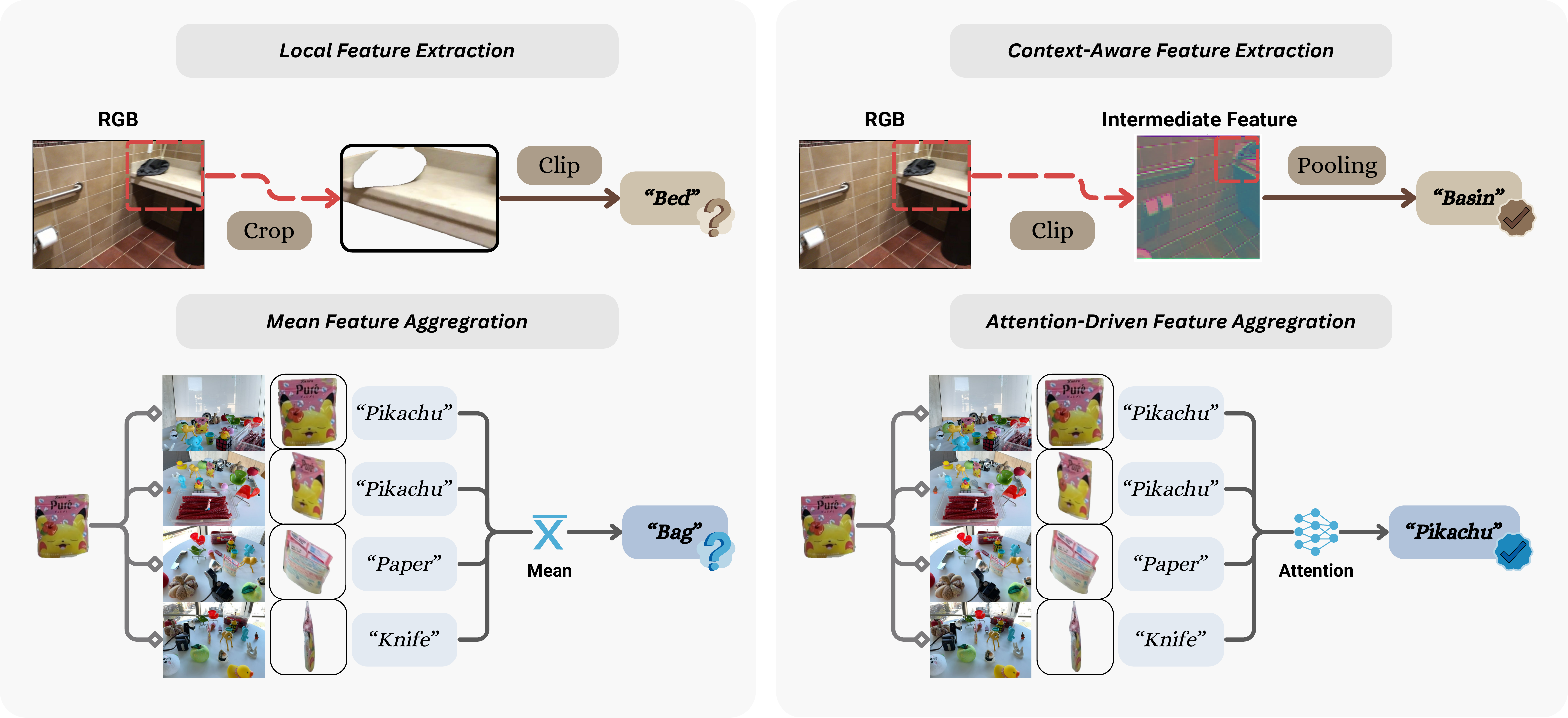}
    \vspace{-0.3cm}
    \captionof{figure}{We introduce OpenInsGaussian, a simple yet effective approach that improves semantic understanding in 3D Gaussian segmentation. By leveraging Context-Aware Feature Extraction and Multi-View Attention, OpenInsGaussian addresses two commonly overlooked issues in existing language understanding pipelines: the loss of contextual information when extracting features from cropped masks and the inconsistencies introduced by multi-view fusion. 
    }
    \label{fig:teaser}
    
    \vspace{0.45cm}
}]



\begin{abstract}
Understanding 3D scenes is pivotal for autonomous driving, robotics, and augmented reality. Recent semantic Gaussian Splatting approaches leverage large-scale 2D vision models to project 2D semantic features onto 3D scenes. However, they suffer from two major limitations: (1) insufficient contextual cues for individual masks during preprocessing and (2) inconsistencies and missing details when fusing multi-view features from these 2D models. In this paper, we introduce \textbf{OpenInsGaussian}, an \textbf{Open}-vocabulary \textbf{Ins}tance \textbf{Gaussian} segmentation framework with Context-aware Cross-view Fusion. Our method consists of two modules: Context-Aware Feature Extraction, which augments each mask with rich semantic context, and Attention-Driven Feature Aggregation, which selectively fuses multi-view features to mitigate alignment errors and incompleteness. Through extensive experiments on benchmark datasets, OpenInsGaussian achieves state-of-the-art results in open-vocabulary 3D Gaussian segmentation, outperforming existing baselines by a large margin. These findings underscore the robustness and generality of our proposed approach, marking a significant step forward in 3D scene understanding and its practical deployment across diverse real-world scenarios.

\end{abstract}
\section{Introduction}
\label{sec:intro}
3D scene understanding has become increasingly crucial for various real-world applications, such as autonomous driving, robotics, and augmented reality. As these domains demand a reliable representation of complex, dynamically changing environments, researchers have turned to 3D Gaussian Splatting (3DGS) \cite{3dgs} for high-fidelity scene modelling. In contrast to traditional point cloud or voxel-based representations, 3DGS encodes a scene as a collection of learnable 3D Gaussian primitives, achieving impressive visual quality and efficient rendering. However, imbuing 3DGS with robust semantic understanding remains a challenging pursuit.

A natural approach to incorporate semantics into 3DGS involves harnessing the power of large-scale 2D vision-language models (VLMs) and projecting 2D VLMs embeddings onto the 3D scene \cite{langsplat, opengaussian, le3dgs, lerf}, thereby augmenting each Gaussian primitive with semantic features. While these works have demonstrated promising results, they face two critical limitations. First, contextual information loss occurs when object segments are cropped for feature extraction, obscuring important cues in the broader image context, especially for small or partially occluded objects. Second, multi-view feature inconsistency arises due to variations in lighting, occlusions, and camera perspectives, causing different views of the same object to yield misaligned embeddings. Simply averaging these features usually propagates noise and degrades overall performance (Fig. \ref{fig:teaser}).

In this paper, we introduce \textbf{OpenInsGaussian}, an open-vocabulary instance Gaussian segmentation framework equipped with a novel Context-aware Cross-view Fusion mechanism to address these limitations. Our approach is built around two key components. First, we propose a Context-Aware Feature Extraction module that pools features directly from the frozen CLIP backbone using mask-based queries, preserving spatial context and leveraging the global semantic knowledge of the model~\cite{maskclip}. This design avoids discarding critical visual cues during preprocessing, enabling richer semantic representations of target objects. Second, we devise an Attention-Driven Feature Aggregation strategy that selectively weights feature contributions from multiple views based on their semantic consistency. By down-weighting noisy or misaligned embeddings, our method produces more accurate and robust 3D semantic reconstructions.

OpenInsGaussian is evaluated on open-vocabulary 3D gaussian segmentation benchmarks and demonstrate that it substantially outperforms prior works, establishing a new state-of-the-art in 3DGS-based instance segmentation. Our experiments further show that incorporating spatial context and adapting attention weights across views significantly enhances the quality and robustness of the reconstructed 3D scenes. This approach marks an important milestone in bridging the gap between high-fidelity 3D representations and semantic understanding, paving the way for broader deployment of 3D vision models in diverse real-world scenarios.

The key contributions of our work are listed as follows:
\begin{itemize}
\item {We introduce a novel 3D Gaussian Splatting approach, OpenInsGaussian, designed to handle open-vocabulary instance-level segmentation by effectively leveraging large-scale vision-language models.}
\item {To address contextual information loss, we propose a mask-based feature extraction strategy that pools features directly from the frozen CLIP backbone.}
\item {We develop an attention-driven feature aggregation module that adaptively weighs multi-view features based on their semantic consistency, mitigating feature inconsistency and enhances overall 3D semantic representations.}
\item {OpenInsGaussian significantly outperforms prior work in both accuracy and robustness, thus setting a new state of the art in 3D instance gaussian segmentation.}
\end{itemize}
\section{Related Work}
\label{sec:related}
\subsection{3D Representation}
\label{subsec:representationd}
3D representation is fundamental for novel view synthesis, 3D reconstruction, and scene understanding. Traditional explicit methods, such as voxels, point clouds, and meshes, directly encode geometry but suffer from high memory usage or lack connectivity. Neural implicit representations, particularly Neural Radiance Fields (NeRF)~\cite{nerf}, have revolutionized novel view synthesis by mapping 3D coordinates to radiance and density values. While NeRF achieves high-quality rendering, its per-scene optimization is computationally expensive, leading to slow training and inference. Efforts to accelerate NeRF include hybrid methods incorporating voxel, tri-plane or hash grids~\cite{dvgo, plenoxels, merf, instantngp}, reducing network complexity while maintaining visual fidelity. 3DGS~\cite{3dgs} further advances neural 3D representation by replacing volume rendering with fast differentiable rasterization. Instead of ray-marching, 3DGS models scenes as a set of learnable 3D Gaussians projected onto the image plane and blended via a splatting operation, enabling real-time rendering while preserving high-quality reconstruction. While 3DGS has been extended to extensive field like dynamic scenes~\cite{dynamic3dgs, deformable3dgs} and generative modeling~\cite{textto3d, dreamgaussian, gaussiandreamer}, these methods focus on visual quality rather than semantic understanding. In contrast, our approach focusing on integrates language embeddings into 3D Gaussian, enabling open-vocabulary 3D point-level understanding and bridging the gap between high-fidelity scene representation and semantic reasoning.

\subsection{3D Class Agnostic Segmentation}
\label{subsec:agnostic_segmentation}

The Segment Anything Model (SAM)~\cite{sam} has demonstrated strong zero-shot segmentation capabilities in 2D images, making it a foundational tool for various computer vision tasks~\cite{yu2023inpaint, tracking, gao2023editanything}. Given its success, recent research has explored extending SAM-based segmentation to 3D~\cite{sa3d, saga, spinnerf, gaussiangrouping, garfield, clickgaussian}. More specifically, research has explored 3D Gaussian Splatting for scene segmentation. SAGA~\cite{saga} applies contrastive learning with SAM-generated masks, using a trainable MLP to project instance features into a low-dimensional space, thereby reducing inconsistencies. Gaussian-Grouping~\cite{gaussiangrouping} leverages a zero-shot tracker~\cite{tracking} to improve mask consistency, jointly reconstruct and segment 3DGS. ClickGaussian~\cite{clickgaussian} introduces Global Feature-Guided Learning, clustering global feature candidates derived from noisy 2D segments across multiple views.

More reviews of existing methods that integrate NeRF~\cite{nerf} with SAM for class-agnostic segmentation is provided in Appendix~\ref{appendix:calss_agnostic_review}.

\subsection{3D Open-Vocabulary Understanding} 
\label{subsec:languagefield}

The emergence of large language models~\cite{brown2020language, bert} has driven rapid advancements in vision-language learning. CLIP~\cite{clip} demonstrated that contrastively pretraining dual-encoder models on large-scale image-text pairs enables cross-modal alignment, achieving strong zero-shot performance on various downstream tasks. 

Integrating 2D vision-language models into 3D representation learning has significantly enhanced open-vocabulary scene understanding. Early efforts in this direction include Distilled Feature Fields~\cite{kobayashi2022decomposing} and Neural Feature Fusion Fields~\cite{tschernezki2022neural}, which distilled multi-view LSeg~\cite{li2022language} or DINO~\cite{caron2021emerging} features into NeRF-based representations. Semantic NeRF~\cite{semanticnerf} further embedded semantic information into NeRFs to enable novel semantic view synthesis, while LERF~\cite{lerf} pioneered the integration of CLIP features within NeRF for open-vocabulary 3D queries. Later approaches, such as 3D-OVS~\cite{liu2023weakly}, combined CLIP and DINO features to improve 3D open-vocabulary segmentation. Nested Neural Feature Fields~\cite{bhalgat2024n2f2} extend feature field distillation by introducing hierarchical supervision that assigns different dimensions of a single high-dimensional field to encode scene properties at multiple granularities, leveraging 2D class-agnostic segmentation and CLIP embeddings for coarse-to-fine open-vocabulary 3D scene understanding.

Despite their effectiveness in leveraging VLMs for 3D semantic reasoning, NeRF-based methods still suffer from computationally expensive volume rendering. To address this limitation, recent studies have explored 3D Gaussian Splatting as a more efficient alternative for real-time neural scene representation. LEGaussians~\cite{le3dgs} introduced uncertainty-aware semantic attributes to 3D Gaussians, aligning rendered semantic maps with quantized CLIP and DINO features. LangSplat~\cite{langsplat} employed a scene-specific language autoencoder to encode object semantics, enhancing feature separability in rendered images. Feature3DGS~\cite{feature3dgs} proposed a parallel N-dimensional Gaussian rasterizer for high-dimensional feature distillation. Unlike LangSplat~\cite{langsplat} and LEGaussians~\cite{le3dgs}, which directly learn multi-view quantized semantic features onto Gaussians, OpenGaussian~\cite{opengaussian} introduces a novel pipeline. It first applies a contrastive learning scheme to generate class agnostic 3D masks, then bound multi-view CLIP features to objects, enabling point-level open-vocabulary understanding.

These advancements highlight the shift from NeRF-based methods to more computationally efficient Gaussian splatting approaches, offering new possibilities for scalable and real-time 3D open-vocabulary scene understanding.

\section{Method}
\label{sec:method}

\subsection{Preliminary: 3D Gaussian Splatting}
\label{sec:3dgs_overview}
A 3D scene is represented as a set of 3D Gaussians, each parameterized by its position $ \mu \in \mathbb{R}^3 $, scale $ S \in \mathbb{R}^3 $, rotation $ R \in \mathbb{R}^4 $, opacity $ \sigma $, and color features $ c $, which are encoded using spherical harmonics (SH). 

For rendering, 3D Gaussian Splatting~\cite{3dgs} follows a splatting-based pipeline, where each Gaussian is projected onto the 2D image space according to the camera’s world-to-frame transformation. Gaussians overlapping at the same pixel coordinates $(x, y)$ are blended in depth order with opacity-weighted accumulation to compute the final pixel color:
\begin{equation}
c_{x,y} = \sum_{i \in N} c_i \alpha_i \prod_{j=1}^{i-1} (1 - \alpha_j),
\end{equation}
where $\alpha_i$ represents the opacity of the $i$-th Gaussian, and $c_i$ is its associated color. This process enables real-time differentiable rasterization, allowing the optimization of Gaussian parameters through a reconstruction loss computed against ground truth images. The flexibility and efficiency of 3D Gaussians have made it a promising approach for high-fidelity 3D reconstruction and novel view synthesis.
\begin{figure*}[t] 
    \centering
    \includegraphics[width=0.95\linewidth]{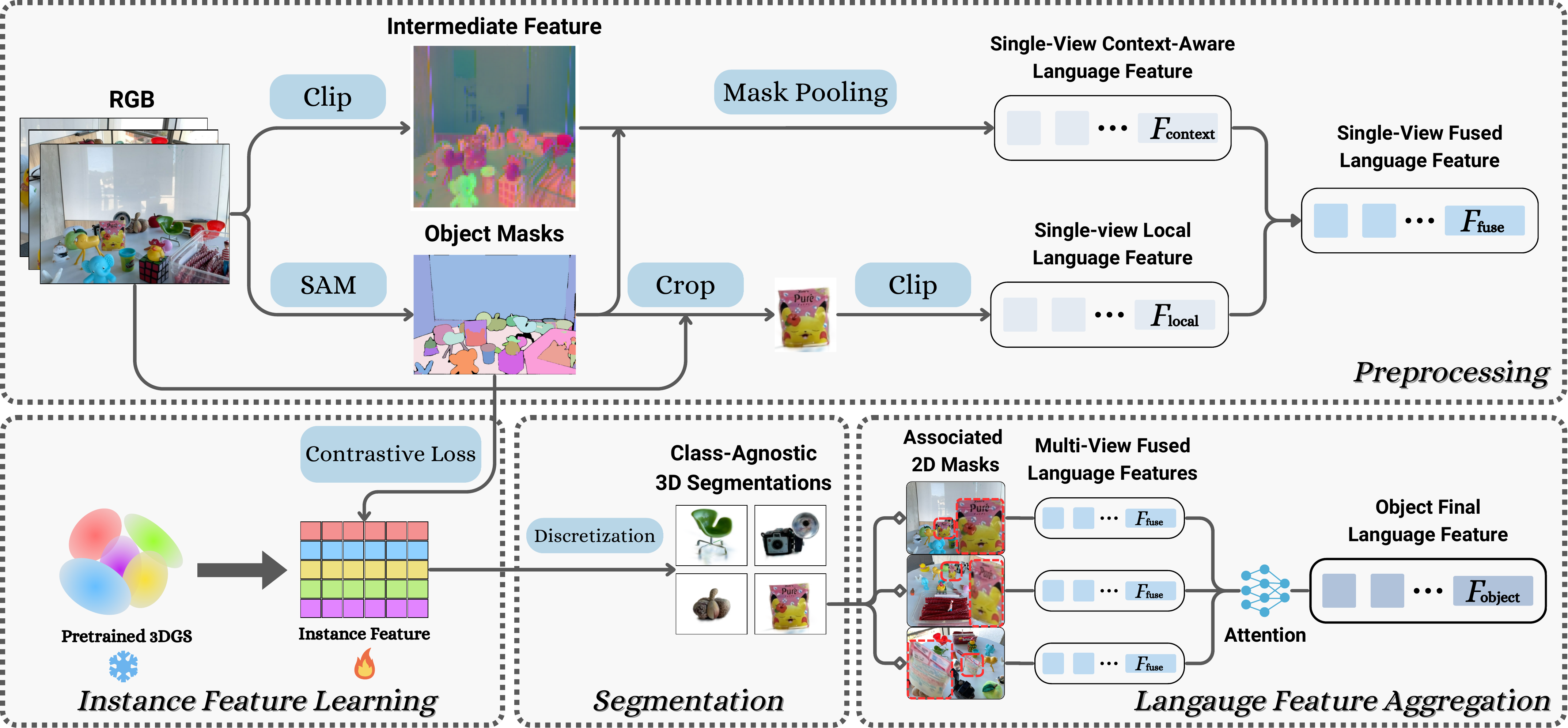}
    \caption{Given a set of images and pretrained 3D Gaussians, our pipeline consists of four key steps: (1) We preprocess the input images using SAM~\cite{saga} to obtain object masks and fuse local and context-aware CLIP~\cite{clip} encoded language features to generate context-aware language embeddings for each mask. (2) The SAM masks are then used to train class-agnostic instance features for each 3D Gaussian. (3) We discretize the 3D Gaussians by clustering instance features using a hierarchical approach. (4) Finally, we employ an attention-driven feature aggregation method to associate language embeddings with the segmented instances, enabling open-vocabulary understanding.}
    \label{fig:overview}
\end{figure*}
\subsection{Method Overview}
\label{sec:method_overview}
Given a set of input images, our approach constructs a 3D Gaussian representation while integrating language embeddings to enable open-vocabulary scene understanding. The objective is to associate each Gaussian with meaningful language features, allowing for robust semantic comprehension. To achieve this, we design a structured training pipeline comprising several key stages.

We begin by initializing a set of pretrained 3D Gaussians and utilizing SAM~\cite{sam} to generate object masks from the input images. To extract robust language features, we introduce a context-aware feature extraction method based on CLIP~\cite{clip} (Sec.~\ref{sec:preprocess}). This method combines fine-grained, distinct local features with context-aware global features, ensuring that each mask is embedded with rich semantic information.

To achieve class-agnostic 3D instance segmentation, we follow OpenGaussian~\cite{opengaussian} to train instance features using SAM masks within a contrastive learning framework. Then discretize the pretrained Gaussians based on these learned instance features using a hierarchical coarse-to-fine clustering strategy (Sec.\ref{sec:class_agnostic}). This process generates class-agnostic 3D masks, forming a structured segmentation of the scene.

Finally, we refine the association between 3D Gaussians and language features by applying a similarity-driven adaptive attention mechanism (Sec.~\ref{sec:attention_aggregation}). This mechanism lifts multi-view 2D language embeddings into the 3D space, ensuring that each Gaussian is effectively aligned with semantic representations. By integrating these components, our method enables robust open-vocabulary understanding within a 3D Gaussian Splatting framework.

\subsection{Context-Aware Feature Extraction}
\label{sec:preprocess}
In this section, we aim to extract mask-level language features from an image batch for open-vocabulary 3D scene understanding. Given a batch of images \( \{I_t \mid t = 1, 2, ..., T\} \), our goal is to extract mask-level language features while preserving contextual information.

Each image \( I_t \in \mathbb{R}^{3 \times H \times W} \) is a standard RGB image, where \( H \) and \( W \) denote the height and width of the image. We use SAM~\cite{sam} to generate instance masks, represented as a mask matrix:
$
    M_t \in \mathbb{R}^{1 \times H \times W},
$
where each pixel value in \( M_t \) indicates the instance label of that region. The set of binary masks for each image is:
$
    \{ B_{t,1}, B_{t,2}, ..., B_{t,N_t} \} = \{ \mathbb{I} (M_t = i) \mid i \in M_t \}.
$
Each binary mask \( B_{t,i} \) is a boolean indicator map where pixels belonging to instance \( i \) are set to 1, and all others are 0.
We extract language embeddings for all masks in a given image through below two complementary methods.

\subsubsection{Local Feature Extraction}
The conventional approach~\cite{langsplat} extracts features by cropping each mask region and passing it through the CLIP image encoder:
\begin{equation}
    F_{\text{local}}(t) = \{V_{\text{clip}}(I_t \odot B_{t,i}) \mid i = 1, 2, ..., N_t\},
\end{equation}
where \( \odot \) denotes element-wise multiplication, effectively cropping out object regions. \( V_{\text{clip}} \) is the entire CLIP image encoding pipeline, which takes an image \( I \) and outputs a global image-level feature in the shared vision-language embedding space.
\begin{figure}[t] 
    \centering
    \includegraphics[width=\columnwidth]{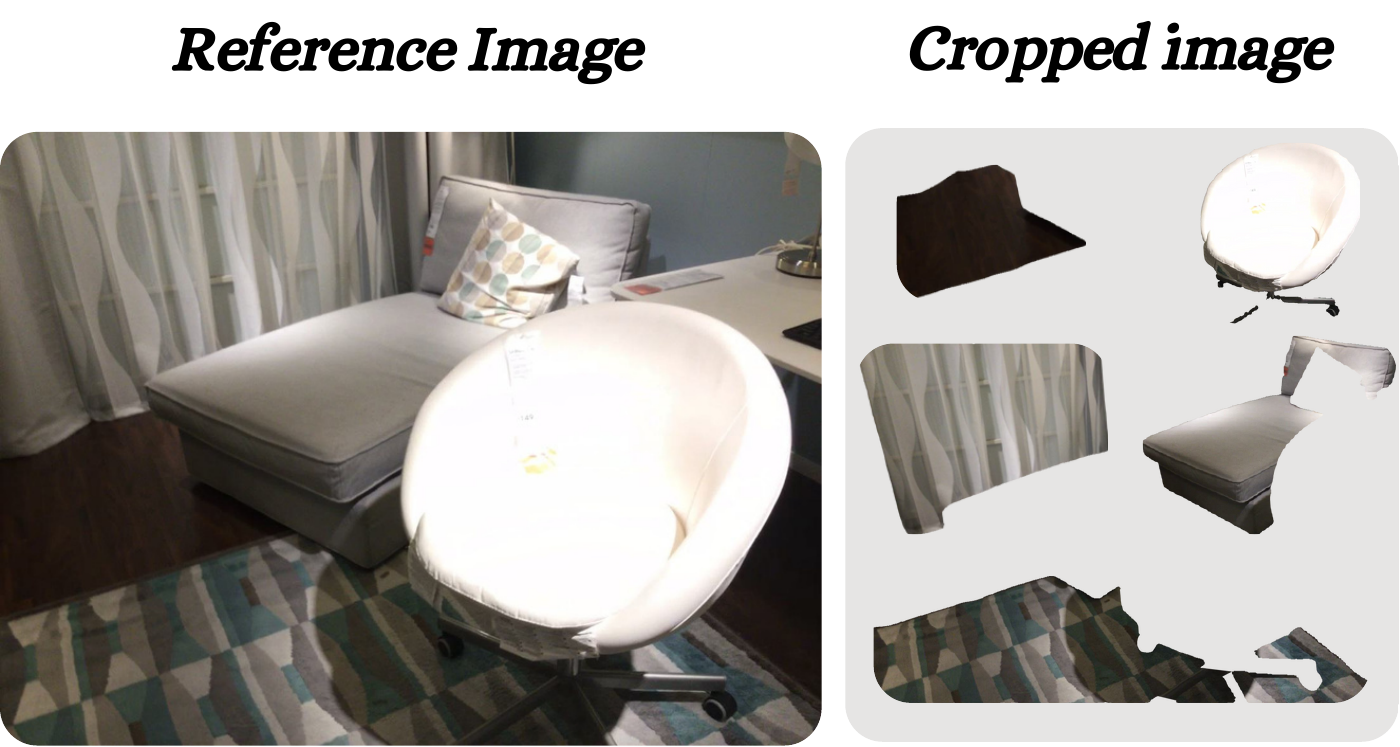}
    \caption{Example of a cropped image patch in local feature extraction on Scannet dataset.}
    \label{fig:crop_image}
\end{figure}
Cropping isolates objects from their surroundings, stripping away crucial spatial context. Many objects, especially in indoor scenes, rely on background information for accurate identification. Also, CLIP is trained to process whole images, and extracting individual object features from cropped patches disrupts the spatial relationships learned in the model. This results in a potential loss of semantic fidelity. Figure~\ref{fig:crop_image} illustrates how cropped patches often lack sufficient contextual cues, leading to ambiguous object recognition. For example a cropped patch of a ``floor" might be indistinguishable without its surrounding environment. 

\subsubsection{Context-Aware Feature Extraction}

CLIP's image encoder processes entire images and extracts spatial feature maps at intermediate layers. However, direct cropping bypasses these feature maps, forcing the model to rely solely on the global image representation. This misalignment can degrade feature quality, particularly for fine-grained open-vocabulary understanding.

To overcome these limitations, we propose a context-aware feature extraction strategy that preserves spatial context while ensuring robust feature alignment. Instead of directly cropping images, we leverage CLIP's intermediate feature maps before the fully connected layer. These feature maps retain rich contextual information about object surroundings, which is crucial for fine-grained open-vocabulary understanding.

Rather than encoding each object independently, we extract spatial feature maps from CLIP's vision encoder and aggregate features within each mask. By cropping the feature maps instead of the raw image, we ensure that object features are learned within the broader scene context rather than in isolation. This approach enables the model to capture finer object details while maintaining surrounding information, leading to more precise and semantically meaningful feature representations. Mathematically, the process define as:
\begin{equation}
    F_{\text{img}}(t) = V_{\text{encoder}}(I_t) \in \mathbb{R}^{D' \times H' \times W'},
\end{equation}
where \( D', H', W' \) are the spatial dimensions of the downsampled feature map, \( V_{\text{encoder}} \) is the convolutional backbone or transformer feature extractor within the CLIP model, which processes an image into a spatial feature map before the final pooling and projection layers

Since the segmentation mask \( M_t \) is at the original image resolution, we resize it to match the feature map size:
\begin{equation}
    M'_t = \text{Resize}(M_t, H', W') \in \mathbb{R}^{1 \times H' \times W'}.
\end{equation}
For each object instance \( i \), we generate a binary mask:
\begin{equation}
    B_{t,i} = \mathbb{I} (M'_t = i) \in \{0,1\}^{H' \times W'}.
\end{equation}
We then perform masked average pooling over the feature map and pass the pooling feature through CLIP's fully connected layer, producing the final image-text latent space feature:
\begin{equation}
    F_{\text{mask}}(t, i) = \frac{\sum_{h,w} B_{t,i}(h,w) F_{\text{img}}(t, :, h, w)}{\sum_{h,w} B_{t,i}(h,w)},
\end{equation}
\begin{equation}
    F_{context} = \{V_{\text{proj}}(F_{\text{mask}}(t, i)) \mid i = 1, 2, ..., N_t\},
\end{equation}
where \( V_{\text{proj}} \) is the final projection layer of CLIP that maps extracted image features into the shared vision-language embedding space.
This ensures that the extracted features retain spatial information while maintaining alignment with the segmentation mask.

\subsubsection{Feature Fusion Strategy}
While our method effectively captures global context, CLIP models are not pre-trained with precise object masks, leading to potential misalignment in extracted features. To mitigate this, we employ a geometric ensemble strategy to fuse local and context-aware CLIP features, ensuring more accurate feature alignment.
\begin{equation}
    F_{\text{fuse}}(t) = \alpha \cdot F_{\text{context}}(t) + (1 - \alpha) \cdot F_{\text{local}}(t),
\end{equation}
where \( F_{\text{context}}(t) \) is derived from context-aware feature extraction, and \( F_{\text{local}}(t) \) comes from local feature extraction. The parameter \( \alpha \in [0,1] \) balances their contributions, \( F_{fuse} \in \mathbb{R}^{D \times N_t} \) represents the extracted language feature embeddings for all masks in an image.
\begin{figure}[t] 
    \centering
    \includegraphics[width=\columnwidth]{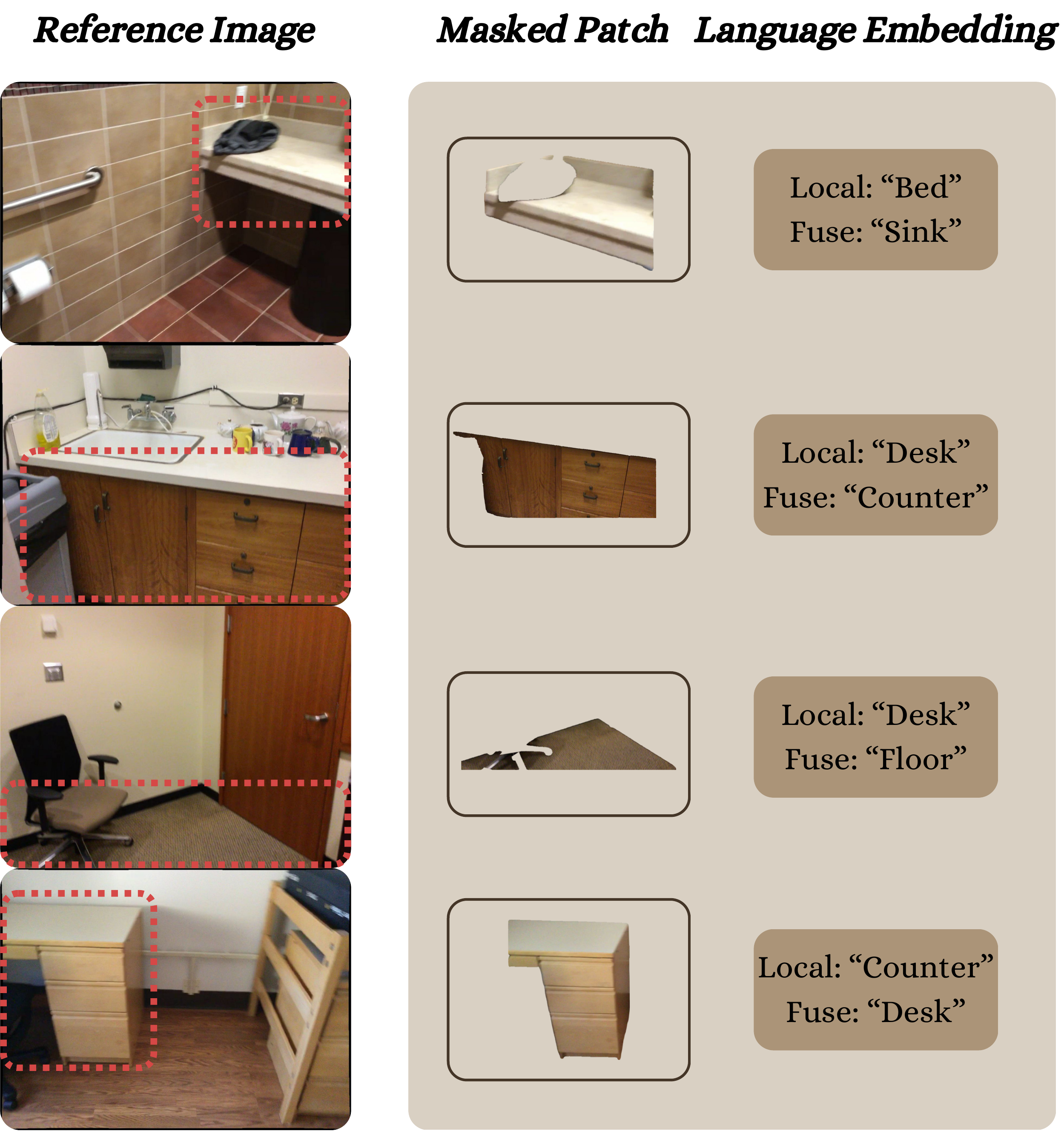}
    \caption{Example of language embedding with different feature extraction strategies on Scannet dataset.}
    \label{fig:fuse_feature}
\end{figure}
By integrating context-aware feature extraction with geometric ensemble fusion, our approach effectively mitigates the challenges posed by missing contextual information and multi-view inconsistencies, leading to a more robust and accurate 3D language field.  Examples in Figure~\ref{fig:fuse_feature} demonstrate that our fused context-aware language features achieve greater accuracy compared to the initial local feature extraction. Here, we present the query text embeddings from the ScanNet test classes, selecting the one with the highest similarity to the masked language features.

\subsection{Class-Agnostic Segmentation}
\label{sec:class_agnostic}
After obtaining segmentation masks from SAM and the corresponding mask-level language embeddings, we follow OpenGaussian~\cite{opengaussian} to segment the scene into class-agnostic clusters. We first train instance features for 3D Gaussians using SAM-generated masks and enforce multi-view consistency through contrastive learning, ensuring that Gaussians within the same mask instance are pulled closer while those from different instances are pushed apart. Next, we apply two-level codebook feature discretization, a coarse-to-fine clustering strategy leveraging both spatial positions and learned instance features to segment objects while preserving geometric integrity. The segmentation is further refined iteratively using pseudo-ground-truth features. Finally, the 3D scene is segmented into class-agnostic clusters. For more details on contrastive learning of instance features and two-level codebook feature discretization, please refer to (Sec.\ref{sec:contrastive_segmentation}, Sec.\ref{sec:feature_discretization}).

\subsection{Attention Driven Feature Aggregation}
\label{sec:attention_aggregation}

After codebook discretization, the scene is segmented into multiple class-agnostic objects. We now binding language features to the objects, started by associating 3d object to multiview 2d masks. The association between 3D Gaussian instances and multi-view 2D masks is established by selecting the highest IoU mask for each rendered 3D instance and refining the assignment using instance feature similarity, ensuring robust alignment. Please refer to (Sec.\ref{aaabbb}) for more details. 

Prior work~\cite{opengaussian} overlooked inconsistencies introduced by varying camera viewpoints. The same object can exhibit significant semantic variation across different views, making simple feature averaging unreliable due to perspective-induced discrepancies. We employ a self-attention-based method for multi-view semantic feature aggregation to associate language embeddings with these segments, ensuring consistent feature binding to each object. Specifically, we propose a similarity-driven attention mechanism for multi-view CLIP feature fusion. Our method adaptively weights features using cosine similarity, enhancing robustness and interpretability while maintaining computational efficiency. Instead of learning query-key interactions, we directly compute cosine similarity between each feature and a reference feature (mean feature) to guide the attention process. This ensures that semantically similar features receive higher weights while inconsistent or occluded views are down-weighted.

Given a set of multi-view CLIP features from section~\ref{sec:preprocess}  $F_{\text{fuse}} = \{F_1, F_2, ..., F_t\}$, we first compute a reference feature as the mean feature across views:
\begin{equation}
F_{\text{mean}} = \frac{1}{N} \sum_{i=1}^{N} F_i.
\end{equation}
Next, we compute the cosine similarity between each feature and the mean feature:
\begin{equation}
S_i = \frac{F_i \cdot F_{\text{mean}}}{\|F_i\| \|F_{\text{mean}}\|}.
\end{equation}
To obtain attention weights, we apply softmax normalization:
\begin{equation}
W_i = \text{softmax}(S_i).
\end{equation}
Finally, the fused feature is computed as a weighted sum:
\begin{equation}
F_{\text{object}} = \sum_{i=1}^{N} W_i F_i.
\end{equation}
where $W_i$ ensures that views with higher semantic consistency contribute more to the final feature representation.

Compared to standard self-attention, our method offers several advantages. It provides robustness to viewpoint variability by explicitly weighting features based on cosine similarity, ensuring that only semantically consistent views contribute significantly, while occluded or noisy views receive lower attention. Also, it improves interpretability, as unlike self-attention, where attention weights are learned implicitly, our approach uses explicit similarity metrics, making it more interpretable. Our method enhances computational efficiency by reducing the complexity from $\mathcal{O}(N^2D)$ in self-attention to $\mathcal{O}(ND)$, making it significantly more scalable. Finally, our approach requires no training, as it is fully unsupervised and can be applied directly to CLIP features without additional fine-tuning. This is particularly useful for scenarios where labeled data is scarce or training a large transformer model is impractical.
\section{Experiment}

\subsection{Open-Vocabulary Query of Point Cloud in 3D Space}

\noindent\textbf{Task.} Given a set of open-vocabulary text queries, the goal of this task is to find the matching point cloud by computing the cosine similarity between the text features and the Gaussian features. Each Gaussian is assigned the category of the text query with the highest similarity, thereby enabling open-vocabulary point cloud understanding. While the method theoretically supports arbitrary text input, for quantitative evaluation against the annotated ground truth point cloud, we use pre-defined category names as text queries.

\noindent\textbf{Dataset and Metrics.} Following OpenGaussian~\cite{opengaussian}, we evaluate our method on ScanNetV2~\cite{scannet}, which provides posed RGB images from video scans, reconstructed point clouds, and ground truth 3D point-level semantic labels. We consider 19, 15, and 10 category subsets from ScanNet as text queries and assign the closest matching text to each Gaussian to compute the mean Intersection over Union (mIoU) and mean Accuracy (mAcc). To ensure consistency, we use the provided point clouds for initialization and freeze the coordinates of the Gaussians during training, disabling the densification process of 3DGS. The training images are extracted every 20 frames from the video scans, and evaluation is conducted on 10 randomly selected scenes.

\noindent\textbf{Baseline.} We primarily compare our method with recent Gaussian-based approaches, including LangSplat~\cite{langsplat}, LEGaussians~\cite{le3dgs}, and OpenGaussian~\cite{opengaussian}.

\begin{figure*}[t] 
    \centering
    \includegraphics[width=0.95\linewidth]{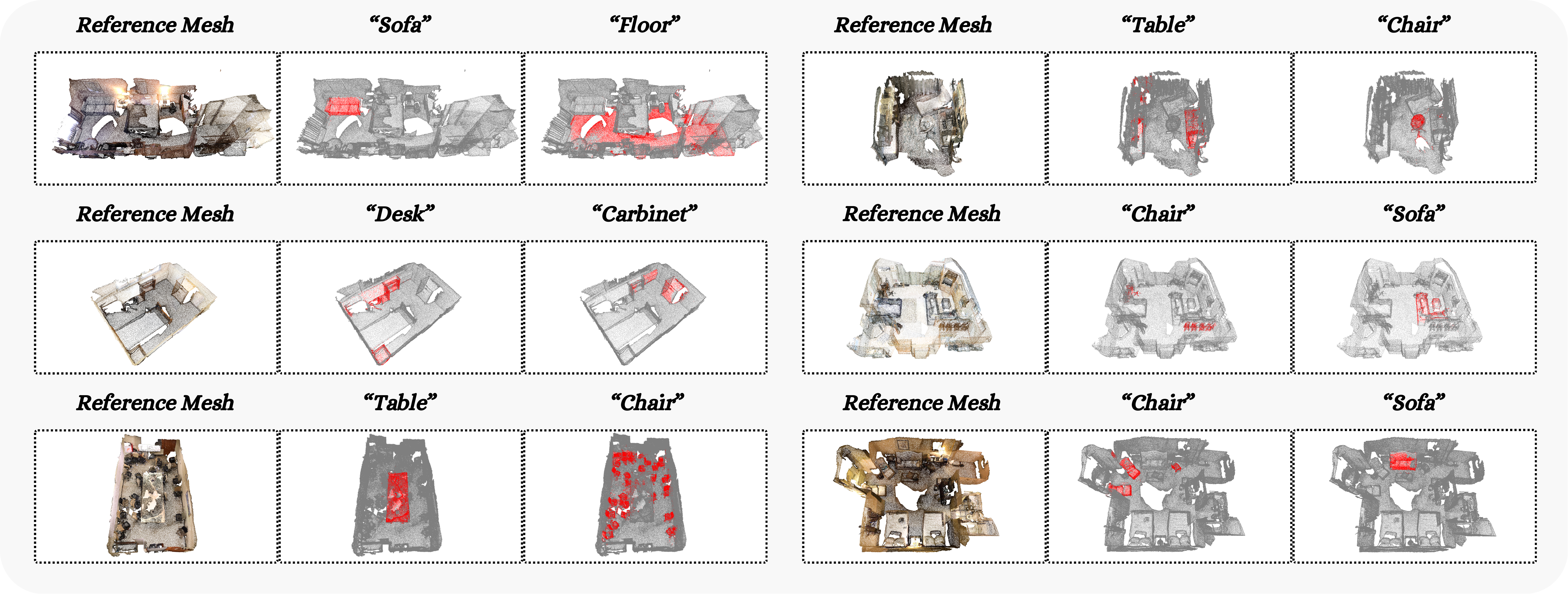}
    \caption{Open-vocabulary 3D point cloud understanding on the Scannet dataset.}
    \label{fig:scannet_example}
\end{figure*}

\noindent\textbf{Results.}
The quantitative results are presented in Table~\ref{tab:scannet_result}, where our method achieves state-of-the-art performance across 19, 15, and 10 categories, significantly outperforming Gaussian-based approaches. The poor performance of LangSplat and LEGaussians can be attributed to ambiguous Gaussian feature learning and suboptimal language feature distillation. OpenGaussian, despite its improvements, is constrained by the nature of the ScanNet dataset, where training images are extracted from video scans. These images often suffer from motion blur, leading to unclear object boundaries and diminished local feature quality. As a result, OpenGaussian accumulates errors in multi-view semantic fusion, further reducing its effectiveness. In contrast, our method leverages context-aware preprocessing and an attention-driven feature fusion module, ensuring more robust semantic binding even under challenging input conditions. Figure~\ref{fig:scannet_example} visualize the effectiveness of our approach.
\begin{table}[t]  
    \centering
    \renewcommand{\arraystretch}{1.1}
    \setlength{\tabcolsep}{3pt}  
    \scriptsize  
    \begin{tabularx}{\columnwidth}{l|c c|c c|c c}
        \toprule
        \multirow{2}{*}{Methods} & \multicolumn{2}{c|}{19 Classes} & \multicolumn{2}{c|}{15 Classes} & \multicolumn{2}{c}{10 Classes} \\
        & mIoU $\uparrow$ & mAcc. $\uparrow$ & mIoU $\uparrow$ & mAcc. $\uparrow$ & mIoU $\uparrow$ & mAcc. $\uparrow$ \\
        \midrule
        LangSplat~\cite{langsplat} & 3.78  & 9.11  & 5.35  & 13.20 & 8.40  & 22.06 \\
        LEGaussians~\cite{le3dgs} & 3.84  & 10.87 & 9.01  & 22.22 & 12.82 & 28.62 \\
        OpenGaussian~\cite{opengaussian} & 24.73 & 41.54 & 30.13 & 48.25 & 38.29 & 55.19 \\
        Ours & \textbf{37.50} & \textbf{54.38} & \textbf{38.14} & \textbf{55.30} & \textbf{51.42} & \textbf{69.15} \\
        \bottomrule
    \end{tabularx}
    \caption{Performance of 3D point cloud semantic segmentation on the ScanNet dataset based on text query.}
    \label{tab:scannet_result}
\end{table}
\subsection{Open-Vocabulary Selection and Rendering in 3D Space}

\noindent\textbf{Task.} Given an open-vocabulary text query, we extract its text feature using CLIP and compute the cosine similarity between this feature and the language features of each Gaussian. Based on the similarity scores, we select the most relevant 3D Gaussians and render them into multi-view 2D images using the rasterization pipeline of 3DGS. This setup enables us to evaluate the effectiveness of our method in selecting 3D objects that match the input text queries.

\noindent\textbf{Dataset and Metrics.} Following LangSplat~\cite{langsplat} and OpenGaussian~\cite{opengaussian}, we conduct experiments on the LeRF-OVS dataset~\cite{lerf}. After rendering the selected 3D objects into 2D images, we compute the mIoU and mAcc by comparing the rendered images with the corresponding ground truth 2D object masks.

\noindent\textbf{Baseline.} Since only Gaussian-based methods possess both 3D point-level perception and rendering capabilities, we primarily compare our method against LangSplat~\cite{langsplat}, LEGaussians~\cite{le3dgs}, and OpenGaussian~\cite{opengaussian}.

\begin{table}[t]
    \centering
    \renewcommand{\arraystretch}{1.1}
    \setlength{\tabcolsep}{4pt}  
    \footnotesize  
    \begin{tabular}{l|cc}
        \toprule
        Methods & mIoU $\uparrow$ & mAcc. $\uparrow$ \\
        \midrule
        LangSplat~\cite{langsplat} & 9.66  & 12.41 \\
        LEGaussians~\cite{le3dgs} & 16.21  & 23.82 \\
        OpenGaussian~\cite{opengaussian} & 38.36  & 51.43 \\
        Ours & \textbf{42.62} & \textbf{62.11} \\
        \bottomrule
    \end{tabular}
    \caption{Performance of open vocabulary 3D object selection and rendering on Lerf dataset. Accuracy is measured by mAcc@0.25}
    \label{tab:lerf_result}
\end{table}

\begin{figure}[t] 
    \centering
    \includegraphics[width=0.95\columnwidth]{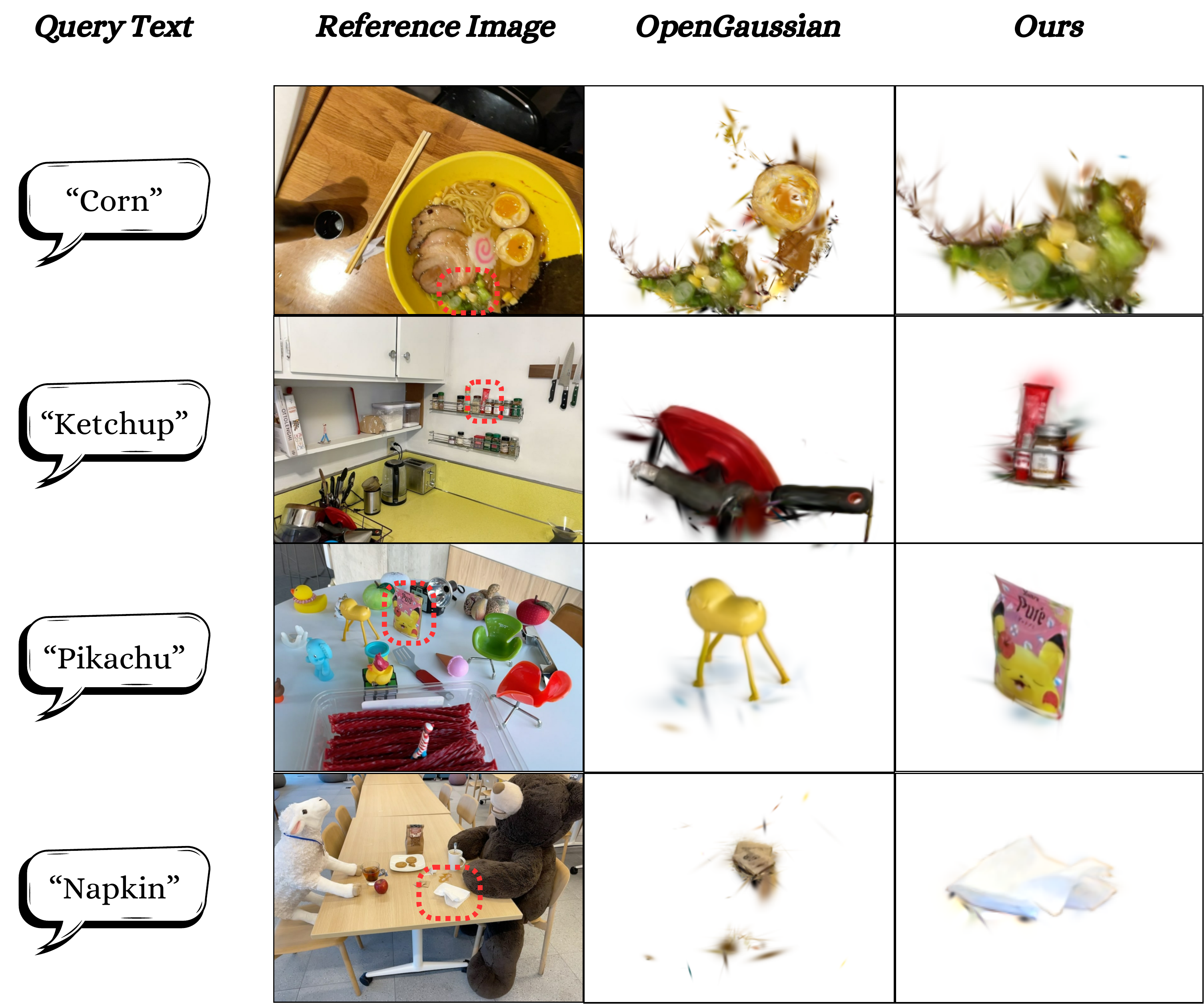} 
    \caption{Open-vocabulary 3D object selection and rendering on the LeRF dataset.}
    \label{fig:lerf_example}
\end{figure}

\noindent\textbf{Results.} 
Quantitative results are presented in Table~\ref{tab:lerf_result}. LangSplat~\cite{langsplat} and LEGaussians~\cite{le3dgs} fail to accurately select 3D Gaussians when rendering text-query-relevant 3D Gaussians onto 2D images for evaluation. Compared to OpenGaussian~\cite{opengaussian}, despite following a similar class-agnostic segmentation process, our method achieves superior performance in language understanding, setting a new state-of-the-art benchmark. The performance advantage of our approach stems from two key aspects. First, the context-aware feature extraction process ensures that each mask retains crucial surrounding information, enhancing semantic consistency. Second, the attention-driven aggregation strategy mitigates the effects of multi-view inconsistencies, a limitation present in OpenGaussian~\cite{opengaussian}. The qualitative results in Figure~\ref{fig:lerf_example} further illustrate these improvements.

\subsection{Ablation Study}
\begin{table}[t]
    \centering
    \renewcommand{\arraystretch}{1.1}
    \setlength{\tabcolsep}{3pt}
    \scriptsize
    \begin{tabularx}{\columnwidth}{c|X|X|X|X}
        \toprule
        Case & Local Feature & Global Feature & mIoU $\uparrow$ & mAcc. $\uparrow$ \\
        \midrule
        \#1 & \checkmark &  & 40.58 & 57.93 \\
        \#2 &  & \checkmark & 50.58 & 67.48 \\
        \#3 & \checkmark & \checkmark & 51.21 & 67.54 \\
        \bottomrule
    \end{tabularx}
    \caption{Performance of 3D point cloud understanding on ScanNet using different feature extraction strategies, evaluated by mIoU and mAcc across 10 classes}
    \label{tab:preprocess_result}
\end{table}

\noindent\textbf{Effect of Context-Aware Feature Extraction} The ablation study results in Table~\ref{tab:preprocess_result} highlight the contributions of local and context-aware feature extraction. Case \#1, using only local features, provides a strong baseline but lacks global context, leading to semantic inaccuracies. Case \#2, relying solely on context-aware features, improves accuracy by preserving scene-wide contextual relationships. Case \#3, which combines both, achieves the best performance, validating our geometric ensemble fusion strategy. These results confirm that local features enhance fine-grained distinctions, while context-aware features provide contextual awareness, making our approach more robust for 3D language field learning.

\begin{table}[t]
    \centering
    \renewcommand{\arraystretch}{1.1}
    \setlength{\tabcolsep}{3pt}
    \scriptsize
    \begin{tabularx}{\columnwidth}{c|X|X|X|X}
        \toprule
        Case & Combined Feature & Attention Aggregation & mIoU $\uparrow$ & mAcc. $\uparrow$ \\
        \midrule
        \#1 &  &  & 40.71 & 57.60 \\
        \#2 & \checkmark &  & 41.98 & 60.52 \\
        \#3 &  & \checkmark & 41.59 & 58.50 \\
        \#4 & \checkmark & \checkmark & 42.62 & 62.11 \\
        \bottomrule
    \end{tabularx}
    \caption{Performance of Open Vocabulary query on LeRF-OVS with Attention-Based Feature Aggregation, evaluated by mIoU and mAcc. Accuracy is measured by mAcc@0.25}
    \label{tab:attention_result}
\end{table}  

\noindent\textbf{Effect of Attention-Driven Feature Aggregation} The results in Table~\ref{tab:attention_result} demonstrate the effectiveness of our attention-driven feature aggregation. Case \#1, which does not use combined features or attention aggregation, provides a baseline performance. Case \#2, which incorporates combined features, shows an improvement in both mIoU and mAcc, indicating that integrating multi-view CLIP features enhances feature robustness. Case \#3, which applies attention aggregation alone, also leads to a performance gain, highlighting the benefits of similarity-driven weighting. Case \#4, where both techniques are combined, achieves the highest scores, confirming that attention-driven fusion effectively reduces inconsistencies in multi-view feature aggregation. These results validate that our proposed method enhances semantic coherence while maintaining fine-grained distinctions, making it a robust approach for 3D open-vocabulary understanding.

\section{Limitation}
\label{sec:limitation}

Since our pipeline segments objects first and then binds language features, we face a bottleneck at the agnostic segmentation stage. Additionally, using the mean as a reference in multi-view attention may be suboptimal when the dominant views contain weak semantic features. Our attention-driven aggregation serves more as a feature purification process, rather than a selection process as commonly used in point cloud-based approaches.

\section{Conclusion}
\label{sec:conclusion}

Lifting 2D large-model semantic features to 3D remains challenging due to inconsistencies in multi-view alignment and semantic drift. Existing methods struggle with feature degradation and ambiguous 2D-3D associations, limiting their effectiveness at the 3D point level. To address this, we introduce a context-aware feature extraction strategy that preserves both local details and context-aware semantics, ensuring richer feature representations. Additionally, we propose attention-driven feature aggregation, leveraging similarity-driven adaptive attention to refine multi-view fusion and improve semantic consistency.  

Our method significantly improves open-vocabulary 3D understanding, achieving state-of-the-art performance across multiple benchmarks. By strengthening the connection between 2D and 3D semantic spaces, we enable more robust and scalable 3D scene interpretation.

\section*{Acknowledgment}
Tongliang Liu is partially supported by the following Australian Research Council projects: FT220100318, DP220102121, LP220100527, LP220200949, and IC190100031.

{\small
\bibliographystyle{ieeenat_fullname}
\bibliography{11_references}
}

\ifarxiv \clearpage \appendix \section{3D Class-agnostic Segmentation}
\subsection{Related Work}
\label{appendix:calss_agnostic_review}
SA3D~\cite{sa3d} uses user-provided prompts, such as points or bounding boxs, to generate segmentation masks in reference views, which are then used to train a neural field for object segmentation. Similarly, Spin-NeRF~\cite{spinnerf} employs a video-based segmenter~\cite{caron2021emerging} to generate multi-view masks. GARField~\cite{garfield} addresses inconsistencies in SAM-generated masks across different views by incorporating a scale-conditioned feature field. OmniSeg3D~\cite{omniseg3d} introduces hierarchical contrastive learning to refine 2D SAM masks into a feature field, achieving fine-grained segmentation through adaptive cosine similarity thresholds. However, above methods rely on NeRF-based structures, which impose high computational costs during rendering, limiting their real-time applicability.

\subsection{Contrastive Feature Learning} 
\label{sec:contrastive_segmentation}
After obtaining segmentation masks from SAM and the corresponding mask-level language embeddings \( \{F_t \mid t = 1, \dots, T\} \), \( \{M_t \mid t = 1, \dots, T\} \), we learn class-agnostic instance features by modeling the relationship between 3D points and 2D pixels. For simplicity, we denote the fused context-aware language feature \( F_{\text{fuse}}(t) \) as \( F_t \) in the following discussion. Following OpenGaussian and other class agnostic segmentation work~\cite{opengaussian, garfield, omniseg3d, clickgaussian, opengaussian}, we train instance features for 3D Gaussians using segmentation masks.

Each Gaussian is assigned a low-dimensional instance feature $f \in \mathbb{R}^6$. To enforce multi-view consistency, we apply contrastive learning, bringing Gaussians within the same mask instance closer while pushing those from different instances apart. The instance feature map $M^f \in \mathbb{R}^{6 \times H \times W}$ is obtained via alpha-blending, with binary masks $B_i$ defining object instances: 
\begin{equation}
    \{ B_0, B_1, \dots, B_i \} = \{ \mathbb{I} (M = i) \mid i \in M_t \},
\end{equation}
To ensure feature consistency within an instance, we compute the mean feature within each mask: \begin{equation} \bar{M}^f_i = \frac{B_i \cdot M^f}{\sum B_i} \in \mathbb{R}^6. \end{equation} The intra-mask smoothing loss encourages all pixels within an instance to align with their mean feature:
\begin{equation}
\mathcal{L}_s = \sum_{i=1}^{m} \sum_{h=1}^{H} \sum_{w=1}^{W} B_{i,h,w} \cdot \left\| M^f_{:,h,w} - \bar{M}^f_i \right\|^2.
\end{equation}
To enhance feature distinctiveness across instances, we define the inter-mask contrastive loss: 
\begin{equation}
\mathcal{L}_c = \frac{1}{m(m+1)} \sum_{i=1}^{m} \sum_{j=1, j \neq i}^{m} \frac{1}{\left\| \bar{M}^f_i - \bar{M}^f_j \right\|^2},
\end{equation} 
where $m$ is the number of masks, and $\bar{M}^f_i, \bar{M}^f_j$ are mean features of different instances.

These losses ensure cross-view consistency for the same object while maintaining feature distinctiveness across different objects.

\subsection{Two-Level Codebook Feature Discretization}  
\label{sec:feature_discretization}  
After training instance features on 3D Gaussians, we apply a two-level coarse-to-fine clustering~\cite{opengaussian} to segment objects.  

At the coarse level, we cluster Gaussians using both 3D coordinates \(X \in \mathbb{R}^{n \times 3}\) and instance features \(f \in \mathbb{R}^{n \times 6}\), ensuring spatially aware segmentation:
\begin{equation}
\begin{split}
f \in \mathbb{R}^{n \times 6}, X \in \mathbb{R}^{n \times 3} \rightarrow
\{C_{\text{coarse}} \in \mathbb{R}^{k_1 \times (6+3)}, \\
I_{\text{coarse}} \in \{1, \dots, k_1\}^n \}, \quad k_1 = 64.
\end{split}
\end{equation}
At the fine level, we further refine clusters using only instance features:
\begin{equation}
\begin{split}
f \in \mathbb{R}^{n \times 6} \rightarrow  
\{C_{\text{fine}} \in \mathbb{R}^{(k_1 \times k_2) \times 6}, \\ 
I_{\text{fine}} \in \{1, \dots, k_2\}^n \}, \quad k_2 = 10.
\end{split}
\end{equation}
where \{C, I\} means quantized features and cluster indices at each level of codebook.

We use K-means clustering~\cite{kmeans} at both stages, with \(k_1\) clusters at the coarse stage and \(k_1 \times k_2\) clusters at the fine stage. This hierarchical approach preserves geometric integrity, ensuring that spatially unrelated objects are not grouped together.  

During instance feature learning, supervision is limited to binary SAM masks. In the codebook construction stage, clustered instance features act as pseudo ground truth, replacing mask-based losses. The new training objective minimizes the difference between rendered pseudo-ground-truth features \(M^p\) and quantized features \(M^c\):
\begin{equation}
\mathcal{L}_p = \| M^p - M^c \|_1,
\end{equation}
This process refines instance segmentation while maintaining feature consistency and geometric structure in the 3D Gaussian representation.

\subsection{Instance-Level 3D-2D Association}
\label{aaabbb}
To establish a robust link between 3D Gaussian instances and multi-view 2D masks, we adopt an instance-level 3D-2D association strategy inspired by OpenGaussian~\cite{opengaussian}. Unlike prior methods that require additional networks for compressing language features or depth-based occlusion testing, our approach retains high-dimensional, lossless linguistic features while ensuring reliable associations.

Specifically, given a set of 3D clusters obtained from the discretization process (Sec.~\ref{sec:feature_discretization}), we render each 3D instance to individual views, obtaining single-instance maps $M^i \in \mathbb{R}^{6 \times H \times W}$. These maps are compared with SAM-generated 2D masks $B^j \in \{0,1\}^{1 \times H \times W}$ using an Intersection over Union (IoU) criterion. The SAM mask with the highest IoU is initially assigned to the corresponding 3D instance. However, to address occlusion-induced ambiguities, we further refine the association by incorporating feature similarity. 

Instead of relying on depth information for occlusion testing, we populate the boolean SAM mask $B^j$ with pseudo-ground truth features, forming a feature-filled mask $P^j \in \mathbb{R}^{6 \times H \times W}$. We then compute a unified association score:

\begin{equation}
    S_{ij} = \text{IoU}(\pi(M^i), B^j) \cdot (1 - \|M^i - P^j\|_1),
\end{equation}

where $\pi(\cdot)$ denotes a binarization operation, ensuring IoU alignment, while the second term penalizes large feature discrepancies. The mask with the highest score is then associated with the 3D instance, allowing us to bind multi-view CLIP features effectively to 3D Gaussian objects.

By integrating both geometric alignment and semantic consistency, our method ensures precise and robust language embedding associations across multiple views.

\section{Implementation Details}
\subsection{SAM and Clip Backbone}
At the preprocessing stage, we utilize SAM-LangSplat, which is a modified version of SAM~\cite{sam} for LangSplat~\cite{langsplat} that automatically generates three levels of masks: whole, part, and sub-part. We select level 3 SAM masks (whole)~\cite{sam} and use the ViT-H SAM model checkpoint for segmentation.

For feature extraction, we adopt Convolutional CLIP~\cite{cnnclip}, a CNN-based variant of CLIP that empirically demonstrates better generalization than ViT-based CLIP~\cite{vitclip} when handling large input resolutions~\cite{fcclip} and better intermediate feature for our global feature extraction. Since competing methods use the ViT-B/16 checkpoint, we select the ConvNeXt-Base checkpoint, which has a comparable ImageNet zero-shot accuracy, ensuring a fair comparison of 2D backbone architectures.

\subsection{Training Strategy}
We follow OpenGaussian~\cite{opengaussian} general training settings. For the ScanNet dataset~\cite{scannet}, that keep point cloud coordinates fixed and disable 3D Gaussian Splatting (3DGS) densification~\cite{3dgs}. For the LeRF dataset~\cite{lerf}, we optimize point cloud coordinates and enable 3DGS densification, which is stopped after 10k training steps.

\subsection{Training Time}
All experiments are conducted on a single NVIDIA RTX 4090 GPU (24GB). For the LeRF dataset, each scene consists of approximately 200 images and requires around 60 minutes for training. For the ScanNet dataset, scenes contain 100–300 images, with an average training time of 30 minutes per scene.

\subsection{ScanNet Dataset Setup}
We align our Scannet Dataset test dataset with ~\cite{opengaussian} on 10 randomly selected ScanNet scenes, specifically:  
\texttt{scene0000\_00}, \texttt{scene0062\_00}, \texttt{scene0070\_00}, \texttt{scene0097\_00}, \texttt{scene0140\_00},  
\texttt{scene0200\_00}, \texttt{scene0347\_00}, \texttt{scene0400\_00}, \texttt{scene0590\_00}, \texttt{scene0645\_00}.  

For text-based queries, we utilize 19 ScanNet-defined categories:  
\begin{itemize}
    \item \textbf{19 categories:} wall, floor, cabinet, bed, chair, sofa, table, door, window, bookshelf, picture, counter, desk, curtain, refrigerator, shower curtain, toilet, sink, bathtub
    \item \textbf{15 categories:} wall, floor, cabinet, bed, chair, sofa, table, door, window, bookshelf, counter, desk, curtain, toilet, sink
    \item \textbf{10 categories:} wall, floor, bed, chair, sofa, table, door, window, bookshelf, toilet
\end{itemize}

Training images are downsampled by a factor of 2, and we use the cleaned point cloud that is processed by OpenGaussian.

\subsection{Addional Results}
\begin{table}[t]
\centering
\caption{Ablation study on ScanNet with different feature aggregation weights. Metrics are reported as mIoU and mAcc for 10, 15, and 19 class settings.}
\resizebox{\columnwidth}{!}{%
\begin{tabular}{ccc}
\toprule
\textbf{Context Feature Weight $\alpha$} & \textbf{mIoU (10, 15, 19)} & \textbf{mAcc (10, 15, 19)} \\
\midrule
0   & 0.42, 0.33, 0.33 & 0.60, 0.50, 0.49 \\
0.2 & 0.51, 0.38, 0.38 & 0.69, 0.55, 0.54 \\
0.4 & 0.51, 0.39, 0.38 & 0.68, 0.56, 0.55 \\
0.6 & 0.50, 0.38, 0.38 & 0.68, 0.56, 0.54 \\
0.8 & 0.47, 0.36, 0.35 & 0.65, 0.54, 0.52 \\
1   & 0.50, 0.37, 0.37 & 0.68, 0.55, 0.53 \\
\bottomrule
\end{tabular}
}
\label{tab:scannet_ablation}
\end{table}

\begin{table*}[t]
  \centering
  \caption{Per-scene performance of 3D point cloud semantic segmentation on the ScanNet dataset based on text query at different class splits (10 / 15 / 19 classes).}
  \label{tab:scannet_scene_result}
  \setlength{\tabcolsep}{6pt} 
  \begin{tabular}{lcccccc}
    \toprule
    \multirow{2}{*}{Scene ID} &
    \multicolumn{2}{c}{10-class} &
    \multicolumn{2}{c}{15-class} &
    \multicolumn{2}{c}{19-class} \\
    \cmidrule(lr){2-3} \cmidrule(lr){4-5} \cmidrule(lr){6-7}
     & mIoU $\uparrow$ & mAcc. $\uparrow$
     & mIoU $\uparrow$ & mAcc. $\uparrow$
     & mIoU $\uparrow$ & mAcc. $\uparrow$ \\
    \midrule
    scene0000\_00 & 0.4744 & 0.7469 & 0.4054 & 0.6208 & 0.4230 & 0.6149 \\
    scene0062\_00 & 0.4103 & 0.6476 & 0.2907 & 0.5372 & 0.2923 & 0.5372 \\
    scene0070\_00 & 0.5227 & 0.6100 & 0.3899 & 0.4497 & 0.3498 & 0.4086 \\
    scene0097\_00 & 0.5607 & 0.7321 & 0.3419 & 0.5689 & 0.3620 & 0.5658 \\
    scene0140\_00 & 0.5781 & 0.7134 & 0.3422 & 0.4249 & 0.2985 & 0.3718 \\
    scene0200\_00 & 0.4767 & 0.6554 & 0.4336 & 0.5452 & 0.4341 & 0.5452 \\
    scene0347\_00 & 0.5587 & 0.6516 & 0.4018 & 0.5599 & 0.4467 & 0.5926 \\
    scene0400\_00 & 0.5169 & 0.6865 & 0.4066 & 0.5902 & 0.4067 & 0.5902 \\
    scene0590\_00 & 0.6052 & 0.7445 & 0.4517 & 0.6253 & 0.3952 & 0.5609 \\
    scene0645\_00 & 0.4388 & 0.7269 & 0.3502 & 0.6075 & 0.3416 & 0.6509 \\
    \midrule
    \textbf{Mean} & \textbf{0.5142} & \textbf{0.6915}
                  & \textbf{0.3814} & \textbf{0.5530}
                  & \textbf{0.3750} & \textbf{0.5438} \\
    \bottomrule
  \end{tabular}
  \label{tab:per_scene_scannet}
\end{table*}
\begin{table}[t]
  \centering
  \caption{Per-scene performance of open vocabulary 3D object selection and rendering on Lerf dataset with mIoU and mAcc at different thresholds.}
  \label{tab:scene_perf_threshold}
  \setlength{\tabcolsep}{4pt}
  \begin{tabular}{lccc}
    \toprule
    Scene & mIoU $\uparrow$ & mAcc@0.25 $\uparrow$ & mAcc@0.5 $\uparrow$ \\
    \midrule
    figurines     & 0.5375 & 0.7679 & 0.5893 \\
    ramen         & 0.2638 & 0.4366 & 0.1972 \\
    teatime       & 0.5855 & 0.7797 & 0.6441 \\
    waldo\_kitchen& 0.3178 & 0.5000 & 0.4091 \\
    \midrule
    \textbf{Mean} & \textbf{0.4262} & \textbf{0.6211} & \textbf{0.4599} \\
    \bottomrule
  \end{tabular}
  \label{tab:per_scene_lerf}
\end{table}
We have conducted an ablation study (Tab.~\ref{tab:scannet_ablation}) to evaluate the impact of different weighting strategies on performance. This analysis demonstrates the robustness of our approach and highlights the sensitivity of the final segmentation quality to the fusion ratio. We also provide per scene evaluation results on Scannet (Tab.~\ref{tab:per_scene_scannet}) and Lerf dataset (Tab.~\ref{tab:per_scene_lerf}).

\subsection{Efficiency}
Regarding inference time, storage memory, feature extraction time, and training memory cost aspects: previous methods like LangSplat\cite{langsplat} and LEGaussians\cite{le3dgs} perform text-query localization by rendering a 2D compressed language embedding map, which is then decoded to match the text query embeddings—this process is slow. In contrast, our approach and OpenGaussian\cite{opengaussian} can directly localize text queries in 3D by searching the codebook. 
Additionally, both LangSplat\cite{langsplat} and LEGaussians\cite{le3dgs} need to maintain the autoencoder decoder network, which requires larger storage memory. Since our context feature extraction only requires one additional forward pass of CLIP, we do not introduce significant additional computation. In our Attention-Driven Feature Aggregation module, we reuse preloaded multi-view features, incurring no extra memory cost. All methods are compatible with the 4090 GPU.
Moreover, in previous methods, the feature extraction process requires running the CLIP model on each segmentation crop at each granularity level, which is very time-consuming. Our context-aware feature extraction module can be downgraded to solely global feature extraction, which significantly improves efficiency while sacrificing very little accuracy, as shown in our ablation results.
Therefore, our proposed method can achieve substantial improvements without a decrease in efficiency.
 \fi

\end{document}